%% file: conference_101719.tex
\documentclass[conference]{IEEEtran}
\IEEEoverridecommandlockouts
\usepackage{cite}
\usepackage{amsmath,amssymb,amsfonts}
\usepackage{graphicx}
\usepackage{textcomp}
\usepackage{xcolor}
\usepackage{algorithm}
\usepackage[]{algpseudocode}
\usepackage{hyperref}
\usepackage{subcaption}
\usepackage{float}
\usepackage{url}
\usepackage{tikz}
\usepackage{multirow}

\usepackage{colortbl}

\def\BibTeX{{\rm B\kern-.05em{\sc i\kern-.025em b}\kern-.08em
    T\kern-.1667em\lower.7ex\hbox{E}\kern-.125emX}}
\begin{document}


\title{MOEA/D with Random Partial Update Strategy}

	\author{\IEEEauthorblockN{Yuri Lavinas\IEEEauthorrefmark{1}, 
	Claus Aranha\IEEEauthorrefmark{1}, 
	Marcelo Ladeira\IEEEauthorrefmark{2} and
	Felipe Campelo\IEEEauthorrefmark{3}}
	\IEEEauthorblockA{\IEEEauthorrefmark{1}University of Tsukuba, Faculty of Engineering, Information and Systems \\Email: lavinas.yuri.xp@alumni.tsukuba.ac.jp;~caranha@cs.tsukuba.ac.jp}
	\IEEEauthorblockA{\IEEEauthorrefmark{2}University of Brasilia, Department of Computer Science\\ Email: mladeira@unb.br}
	\IEEEauthorblockA{\IEEEauthorrefmark{3}Aston University, School of Engineering and Applied Sciences\\ Email: f.campelo@aston.ac.uk}
	}

\maketitle

\begin{abstract}
 \input{files/0_abstract.tex}
\end{abstract}

\begin{IEEEkeywords}
Multi-Objective Optimization, MOEA/D, Resource Allocation, Partial Update Strategy
\end{IEEEkeywords}

\input{files/1_introduction.tex}

\input{files/2_background.tex}
\input{files/3_methods.tex}

\input{files/4_parameter_study.tex}

\input{files/5_comparision_study.tex}
\input{files/6_conclusion.tex}

\bibliographystyle{IEEEtran}
\bibliography{samplebib}

\end{document}

%% file: files/0_abstract.tex
Recent studies on resource allocation suggest that some subproblems are more important than others in the context of the MOEA/D, and that focusing on the most relevant ones can consistently improve the performance of that algorithm. These studies share the common characteristic of updating only a fraction of the population at any given iteration of the algorithm. In this work we investigate a new, simpler partial update strategy, in which a random subset of solutions is selected at every iteration. The performance of the MOEA/D using this new resource allocation approach is compared experimentally against that of the standard MOEA/D-DE and the MOEA/D with relative improvement-based resource allocation. The results indicate that using the MOEA/D with this new partial update strategy results in improved HV and IGD values, and a much higher proportion of non-dominated solutions, particularly as the number of updated solutions at every iteration is reduced.

%% file: files/1_introduction.tex
\section{Introduction}~\label{intro}

Multi-objective Optimization Problems (MOPs) appear in many application contexts in which several conflicting objective functions need to be simultaneously optimized. Finding good sets of solutions for general continuous MOPs, particularly when convexity or differentiability cannot be assumed, is generally considered a hard problem, for
which Evolutionary Algorithms have been proposed as potential solvers \cite{trivedi2017survey,li2009multiobjective,zitzler2004indicator}.

The Multi-Objective Evolutionary Algorithm Based on Decomposition (MOEA/D)~\cite{zhang2007moea} is generally considered an effective algorithm for solving MOPs. The key idea of the MOEA/D is to decompose the multi-objective optimization problem into a set of single-objective subproblems, which are solved simultaneously by a population-based evolutionary approach. While the original MOEA/D and some of its earlier variants did not discriminate between subproblems, it has since become clear that focusing computational effort on certain subsets of these subproblems can substantially improve the performance of the algorithm \cite{zhang2009performance,zhou2016all,kang2018collaborative,lavinas2019using,pruvost2020subproblemselection}. It has been noted that the MOEA/D may sometimes waste computational effort by trying to improve solutions that are not very promising~\cite{bezerra2015comparing}. This can be a critical issue, particularly  in certain applied MOPs which require costly simulations to evaluate solutions~\cite{kohira2018proposal}. 

To address this issue, a number of works have proposed and investigated methods to allocate different amounts of computational effort to subproblems, based on a variety of \textit{priority functions}~\cite{zhang2009performance, zhou2016all, lavinas2019using, lavinas2019improving, wang2019new}. These approaches, which became collectively known as Resource Allocation (RA) techniques, have been shown to result in consistent performance improvements for the MOEA/D.

While different RA techniques have their specificities, all share the common feature of limiting the number of solutions from the population that are updated at any iteration. On a previous work~\cite{lavinas2019improving} we observed the somewhat surprising result that assigning random priority values to subproblems performed better than not using RA at all. Interestingly enough, Pruvost et. al~\cite{pruvost2020subproblemselection} also found that selecting a subset of subproblems at random on MOEA/D performs well on combinatorial domain. This suggests that increasing the inertia of the population dynamics in the MOEA/D can be beneficial in itself, regardless of the Resource Allocation strategy.

The question that consequently arises can be summarized as: how much of the performance improvements observed in MOEA/Ds with Resource Allocation is due to the RA strategy itself, and how much can be attributed to the simple increase in the inertia of the population dynamics of the MOEA/D, which results from maintaining parts of the population unchanged between iterations? This work focuses on the question of investigating and quantifying the extent of these effects. We also analyse which proportion of the population should be updated at any given iteration, so as to obtain improved performances for the MOEA/D. To investigate these questions we introduce a Partial Update strategy, which allows us to control the proportion of the subproblems that are selected for variation at any iteration. We perform an experimental investigation of the effect of a proportion parameter $ps$ on the performance of the MOEA/D on standard problem benchmarks.

The remainder of this paper is organized as follows: Section \ref{section:RA} provides a brief review of the main concepts related to resource allocation in the MOEA/D. Section \ref{chap3:moead_ps_explanation} introduces the MOEA/D with Partial Update strategy. Sections \ref{partial_update_strategy} and \ref{section:comparison} present experimental results related to the investigation of the effect of partial updates on the performance of the algorithm, as well as comparisons against a baseline algorithm and an existing MOEA/D with resource allocation, using the Relative Improvement priority function. Finally, section \ref{section:conclusion} presents our concluding remarks.

%% file: files/2_background.tex
\section{Resource Allocation}
\label{section:RA}

The key idea behind MOEA/D is to decompose a MOP into a set of single-objective subproblems, which are then solved simultaneously. While these subproblems are usually considered equivalent, a growing body of work has come to indicate that prioritizing some subproblems at certain points of the search can improve the performance of MOEA/D. This issue is commonly addressed by means of resource allocation (RA) techniques. 

Priority functions are used in resource allocation to determine preferences between
subproblems. These functions take information about the 
progress of the search, and return priority values that are then used to change the distribution of computational resources among subproblems at any given iteration~\cite{cai2015external}. This also allows the design of MOEA/D variants that allocate more resources on any desired solution characteristics~\cite{lavinas2019improving}, such as  diversity or robustness \cite{goulart2017robust}.


%



The distribution of computational resources based on priority functions is mediated by a thresholding operation. At any given iteration $t$, let $u_i^t$ indicate the priority function value attributed to the $i$-th subproblem, and $\upsilon^t$ be a threshold value. The subset of solutions selected for variation on that iteration is then defined as the subproblems for which $u_i^t\geq \upsilon^t$.

The original work on resource allocation for the MOEA/D \cite{zhang2009performance} defined a priority function known as the Relative Improvement (RI), defined as:

\begin{equation}\label{sec2:ri}
	u_i^t = \dfrac{f\left(\mathbf{x}_i^{t-\Delta t}
	\right)-f\left(\mathbf{x}_i^t\right)}{f\left(\mathbf{x}_i^{t-\Delta t}\right)},
\end{equation}

\noindent where $f\left(\mathbf{x}_i^{t}\right)$ represents the aggregation function value of the incumbent solution to the $i$-th subproblem on iteration $t$, and $\Delta t$ is a parameter that controls how many generations to wait for the relative improvement comparison (notice that this definition assumes a minimisation problem and an aggregation function that always yields strictly positive values). 

\subsection{State of the art}
Much of the research on resource allocation has used RI as a priority function, with some modifications on other aspects of the algorithm \cite{nasir2011improved,zhou2016all}. \textit{Zhou et al.} did expand the discussion over resource allocation in their work \cite{zhou2016all}, but few other works have studied resource allocation in depth. 

For example, both MOEA/D-GRA and MOEA/D-DRA use the RI priority function. That said, MOEA/D-GRA \cite{nasir2011improved} uses a different replacement strategy to avoid newly generated solutions from updating several neighboring subproblems at any iteration, while MOEA/D-DRA \cite{zhang2009performance} performs a more complex strategy for selecting subproblems using a 10-tournament selection based on the RI priority values. 

Two works that attempted to investigate distinct priority functions are the EAG-MOEA/D~\cite{cai2015external} and MOEA/D-CRA~\cite{kang2018collaborative}. Both used priority functions which allocate resources according to the possibility that subproblems may either be improved or contribute to the improvement of other subproblems. 

In previous works we isolated the priority function as a point of investigation~\cite{lavinas2019using, lavinas2019improving}. The goal in these works was to improve the performance of MOEA/D based on the choice of priority function, and to further understand the behavior of MOEA/D under different resource allocation approaches. On these works, we introduced two new priority functions (DS and iDS), based on the conjecture that MOEA/D would benefit from a greater focus on diversity in the space of objectives. Experimental comparisons were performed between the MOEA/D under three priority functions: (1) RI, (2) DS and iDS; and  under two methods used as baseline: (1) MOEA/D using randomly assigned priority values and (2) MOEA/D without any resource allocation method. 

These experiments revealed the somewhat surprising result that using a random resource allocation performed as well as RI, and better than not using resource allocation at all. This suggests that MOEA/D may benefit simply from the increased populational inertia (possibly due to  slower diversity loss) that results from holding portions of the population constant during any given iteration.

To further investigate this question we propose using an update strategy for the MOEA/D based on randomly allocated priority values. This update strategy allows us to control the expected number of subproblems modified at any given iteration and, consequently, to (partially) regulate the population dynamics of the MOEA/D. This approach is described in the next section.





%% file: files/3_methods.tex
\section{A New Update Strategy for MOEA/D}\label{chap3:moead_ps_explanation}

To verify whether there is a positive effect in limiting the number of solutions that are updated at each iteration, and also to investigate the extent of this effect, we introduce a \textit{Partial Update strategy}. This strategy is used to define the expected amount of solutions updated at each iteration, regulated by a control parameter, $ps\in\left(0,1\right]$. This parameter represents the probability that a given subproblem will be selected for updating at a given iteration. Notice that, under this definition, the allocation of resources to subproblems is completely random, and any effects observed on the performance of the MOEA/D under this allocation strategy will be due only to the effect of maintaining portions of the population unchanged across iterations and their influence with each other. Algorithm~\ref{algo:moead_ps} details the pseudocode of the MOEA/D using the Partial Update Strategy.




\input{files/moead_ps.tex}

Notice that the standard MOEA/D, as well as variants such as MOEA/D-DE \cite{zhang2009performance}, can be instantiated from Algorithm \ref{algo:moead_ps} by setting $ps = 1$. The only difference that the partial update strategy introduces in the base algorithm is that only a few subproblems are updated (probabilistically) at any given iteration, regulated by the value of $ps$. 

Also, MOEA/D-PS maintains the $\Delta t$ parameter from RI. Since MOEA/D-PS does not have an explicit priority function, this parameter just makes the algorithm work in two phases during the search progress. In the first phase all subproblems are updated at every iteration, i.e., with no difference from the usual MOEA/D approach. This initial phase lasts for $\Delta t$ iterations. After that, the algorithm moves onto the second phase, during which MOEA/D-PS performs (randomly selected) partial updates at every iteration.

With this structure other resource allocation techniques could also be expressed, by modifying the priority value attribution function in Line \ref{util_line} of the algorithm (and possibly setting $\Delta t$ to zero, if the initial phase is not desired).

It is important to observe that subproblems that are not selected by the partial update strategy at a given iteration may still have their incumbent solutions updated. Resource allocation in MOEA/D-PS affects only the variation step, not the replacement one; thus, subproblems not selected for variation may receive new candidate solutions, e.g. generated for a neighboring subproblem.

%% file: files/moead_ps.tex
\begin{algorithm}[htbp]
	\caption{MOEA/D-PS (MOEA/D with Partial Update Strategy)}\label{algo:moead_ps}
	\begin{algorithmic}[1]
		\State \textbf{Input}: $ps$, $\Delta t$, Termination criteria, MOEA/D parameters.
		\State \textbf{Initialize} MOEA/D variables (e.g. weight vectors, set of solutions, etc.)
		\State $t \leftarrow 0$
		\State ${u_i} \leftarrow 1$
		\While{\textit{Termination criteria}}
		    \State $t \leftarrow t + 1$
		    \If{$t \geq \Delta t$} \label{algo:phase_2}
		        \State ${u_i} \leftarrow ps$\label{util_line} \Comment{Allocation of update probability}
		    \EndIf
		    \For {i = 1 to N} \Comment{Number of subproblems}
		        \If{\text{rand()} $< u_i$}
		            \State \textbf{Generate} new candidate $y$ for subproblem $i$.
	            \EndIf
	            \State \textbf{Update} the set of solutions by $y$.
		    \EndFor
		    \State  \textbf{Evaluate} the set of solutions.
		\EndWhile
	\end{algorithmic}
\end{algorithm}

%% file: files/4_parameter_study.tex
\section{Partial Update Strategy Parameter Study}\label{partial_update_strategy}

To isolate and examine the effects of updating only part of the MOEA/D population at any iteration, we performed a comparative experiment using two known benchmark sets. Six update levels were used, with $ps\in\left\{0.1, 0.2, 0.4, 0.6, 0.8, 1.0\right\}$ (the last of which simply selects all subproblems for update at every iteration, and represents the standard algorithm without any resource allocation strategy). The MOEA/D-DE implemention from the MOEADr package~\cite{moeadr_package,moeadr_paper} was used as a basis, with modifications included to enable the use of the Partial Update technique as described in the previous section. 
Note that when the $ps$ parameter is is equal to $1.0$, all subproblems are selected to be updated, therefore this case simply reproduces the standard MOEA/D-DE. 

\subsection{Test Problems}
Two benchmark sets were used: the scalable DTLZ set~\cite{deb2005scalable}, with 2 objectives,  and the UF set~\cite{zhang2008multiobjective}. In both cases we used the test functions with dimension $D = 100$. The implementation of the test problems available from the \textit{smoof} package \cite{smoof} was used in all experiments.

The DTLZ suite is composed of seven unconstrained test problems, with distinct problem features~\cite{huband2006review}:

\begin{itemize}
	\item DTLZ1: Linear Pareto Front - unimodal;
	\item DTLZ2: Concave Pareto Front - unimodal;
	\item DTLZ3: Concave Pareto Front - multimodal;
	\item DTLZ4: Concave Pareto Front - unimodal;
	\item DTLZ5: Degenerate Pareto Front - unimodal;
	\item DTLZ6: Degenerate Pareto Front - unimodal;
	\item DTLZ7: Disconnected Pareto Front with concave and convex portions - multimodal.
\end{itemize}

The UF test problems is composed of ten unconstrained test problems with Pareto sets that are designed to be challenging to existing algorithms~\cite{li2019comparison}. Problems UF1-UF7 represent two-objective MOPs, while UF8-UF10 are three-objective problems~\cite{zhang2008multiobjective}.

\begin{itemize}
	\item UF1: Convex Pareto Front - multimodal;
	\item UF2: Convex Pareto Front - multimodal;
	\item UF3: Convex Pareto Front - multimodal;
	\item UF4: Concave Pareto Front - multimodal;
	\item UF5: Linear Pareto Front - multimodal;
	\item UF6: Linear Pareto Front - multimodal;
	\item UF7: Linear Pareto Front - multimodal;
	\item UF8: Concave Pareto Front - multimodal;
	\item UF9: Linear and discontinuous Pareto Front - multimodal;
	\item UF10: Concave Pareto Front - multimodal;
\end{itemize}

\subsection{Experimental Parameters}~\label{parameters}

We used the MOEA/D-DE parameters as they were introduced in the work of Li and Zhang~\cite{li2009multiobjective} in all tests. Table~\ref{chap4:parameter_table} summarizes the experimental parameters. Regarding the $\Delta t$ parameter, we use the value suggested by Zhang when RI was introduced~\cite{zhang2009performance}. We make these parameter 
choices to isolate the contribution of the $ps$ parameter which controls the proportion
of subproblems updated, comparing this change directly with the original algorithms.
Details of these parameters can be found in the documentation of package MOEADr, as well as in the original MOEA/D-DE reference~\cite{zhang2009performance,moeadr_package,moeadr_paper}. All objectives were linearly scaled at every iteration to the interval $\left[0,1\right]$, and the Weighted Tchebycheff scalarization function was used.


\input{files/parameter_table.tex}

\begin{figure*}[htbp]
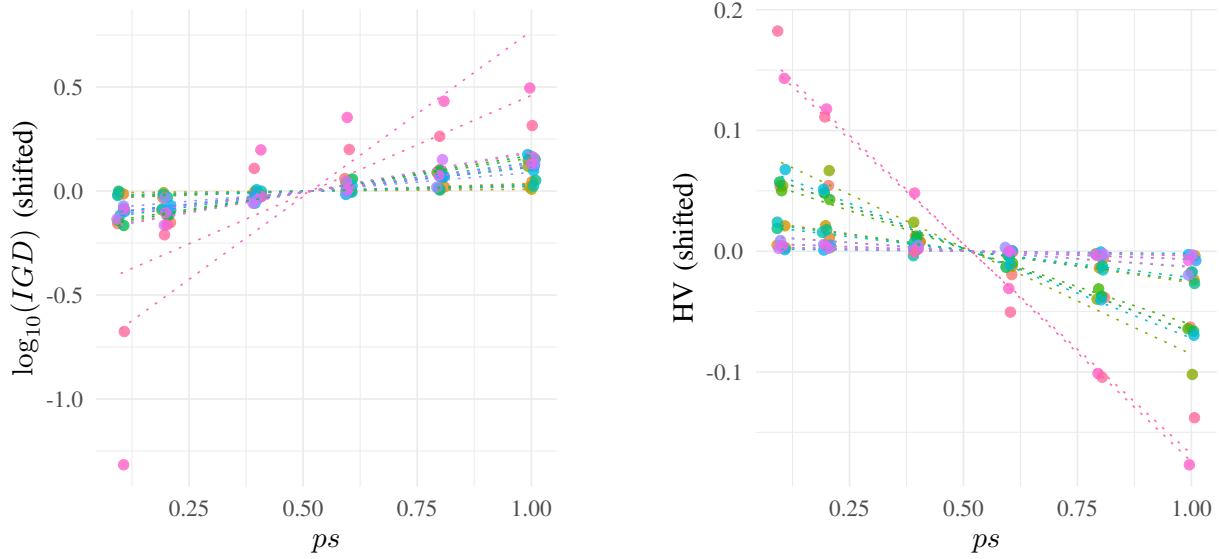

\centering
    \begin{subfigure}[!t]{0.45\textwidth}
        	\input{images/igd_res.tex}
	\end{subfigure}
	~~~
	\centering
    \begin{subfigure}[!t]{0.45\textwidth}
        	\input{images/hv_res.tex}
	\end{subfigure}
	\caption{Linear regression of IGD (lower is better, left) and HV (higher is better, right) against the $ps$ parameter values. Each line represents an individual problem. 
	It is clear that lower values of $ps$ (smaller proportion of updated subproblems) are associated with better performance. }
    	\label{fig:linear_regression}
\end{figure*}

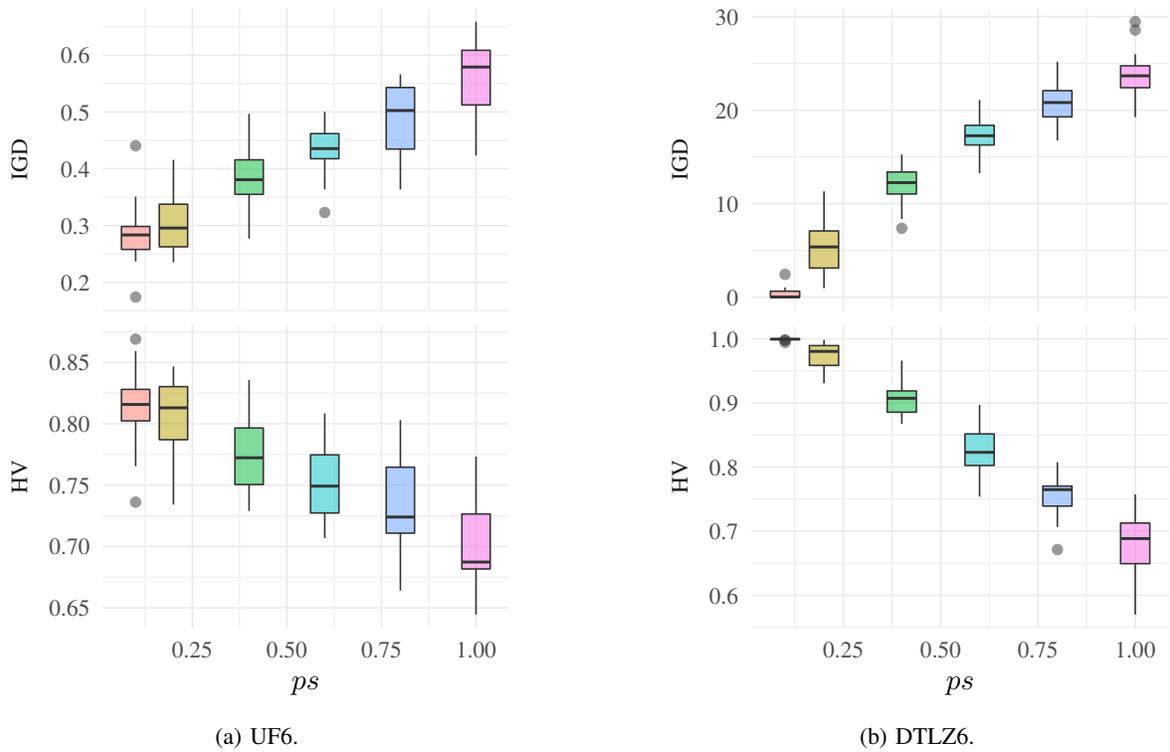
\begin{figure*}[htbp]
\centering
    \begin{subfigure}[!t]{0.45\textwidth}
        	\input{images/box_UF6.tex}
        	\caption{UF6.}
	\end{subfigure}
	~~~
    \begin{subfigure}[!t]{0.45\textwidth}
        	\input{images/box_DTLZ6.tex}
        	\caption{DTLZ6.}
	\end{subfigure}
	\caption{Examples of distribution of IGD values (lower is better) and HV values (higher is better) for the final population, according to $ps$. Problems UF6 on the left and DTLZ6 on the right.}
    	\label{fig:boxplots}
\end{figure*}

\subsection{Experimental Evaluation}~\label{evaluation}

We compare the results of the different strategies using the Hypervolume (HV, higher is better) and Inverted Generational Distance (IGD, lower is better) indicators. We also evaluate the proportion of non-dominated solutions in the final population. The differences among the techniques are analysed using Wilcoxon Rank Sum Tests (all-vs-all), with a  significance level $\alpha = 0.05$ and Hommel adjustment for multiple comparisons. For reproducibility purposes, all the code and experimental scripts are available online~\footnote{\label{gitnote}\href{https://github.com/yclavinas/MOEADr/tree/cec2020}{https://github.com/yclavinas/MOEADr/tree/cec2020}}.

For the calculation of HV, the objective function was scaled to the $(0,1)$ interval, with reference points set to $(1,1)$, for two-objective problems; and $(1,1,1)$, for three-objective ones. 

\subsection{Results}~\label{igd_results}
Figure~\ref{fig:linear_regression} shows regression lines of performance on $ps$ for each test problem, both for log-IGD and HV (higher is better).\footnote{The log transformation was used to account for the large differences of scale in the IGD indicator due to DTLZ1 and DTLZ3, in which all configurations failed to adequately converge. It is possible that the computational budget of $30,000$ candidate solution evaluations may not be enough for solving these two problems.}  These results suggest a clear association between lower values of $ps$ and improved performance on both indicators. Figure~\ref{fig:boxplots} provides a closer visualization of this effect in the case of two test problems, UF6 and DTLZ6 respectively. Statistical tests corroborate these observations, as summarised in Table~\ref{chap4:pvals}. The final raw data and analysis scripts can be retrieved from the project repository on Github.$^1$

\input{files/stats.tex}


\subsection{Anytime Performance of MOEA/D with Partial Update Strategy}\label{chap4:anytime}

Besides providing good final results, it is often desired that a MOEA be capable of returning a set of reasonably good solutions if interrupted at any time during the search~\cite{tanabe2017benchmarking,radulescu2013automatically}. To investigate the impact of using distinct $ps$ values on the anytime performance of the MOEA/D with Partial Update we analyzed the effects in terms of both IGD and HV values.

\begin{figure*}[htbp]
\centering
    \begin{subfigure}[!t]{0.45\textwidth}
        	\includegraphics[width=1\textwidth]{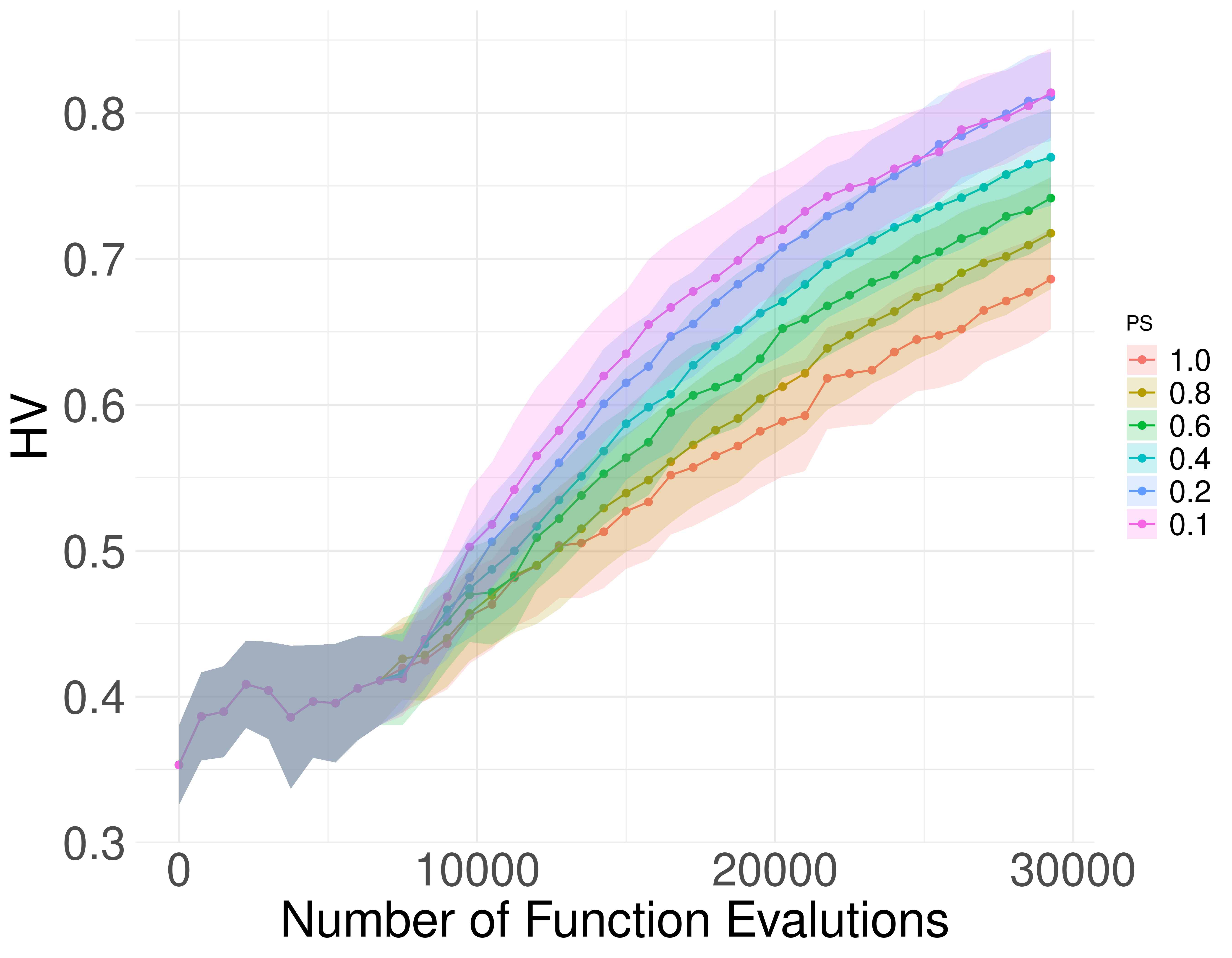}
        	\caption{UF6.}
	\end{subfigure}
	~~
    \begin{subfigure}[!t]{0.45\textwidth}
        	\includegraphics[width=1\textwidth]{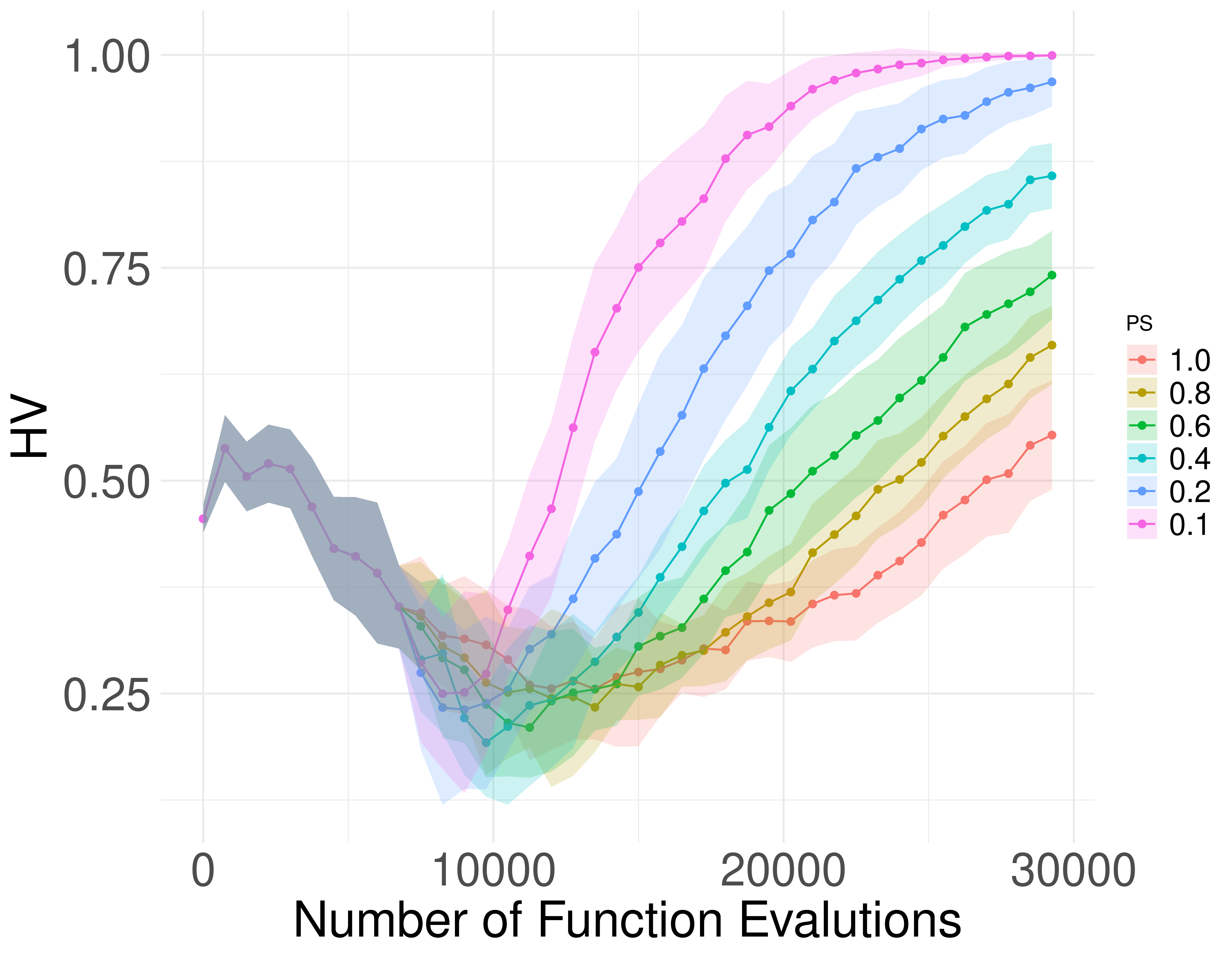}
        	\caption{DTLZ6.}
	\end{subfigure}
	\caption{Anytime HV (higher is better) performance of MOEA/D-PS for different values of $ps$ on two functions.}
	\label{fig:anytime_hv}
\end{figure*}
\begin{figure*}[htbp]
\centering
    \begin{subfigure}[!t]{0.45\textwidth}
        	\includegraphics[width=1\textwidth]{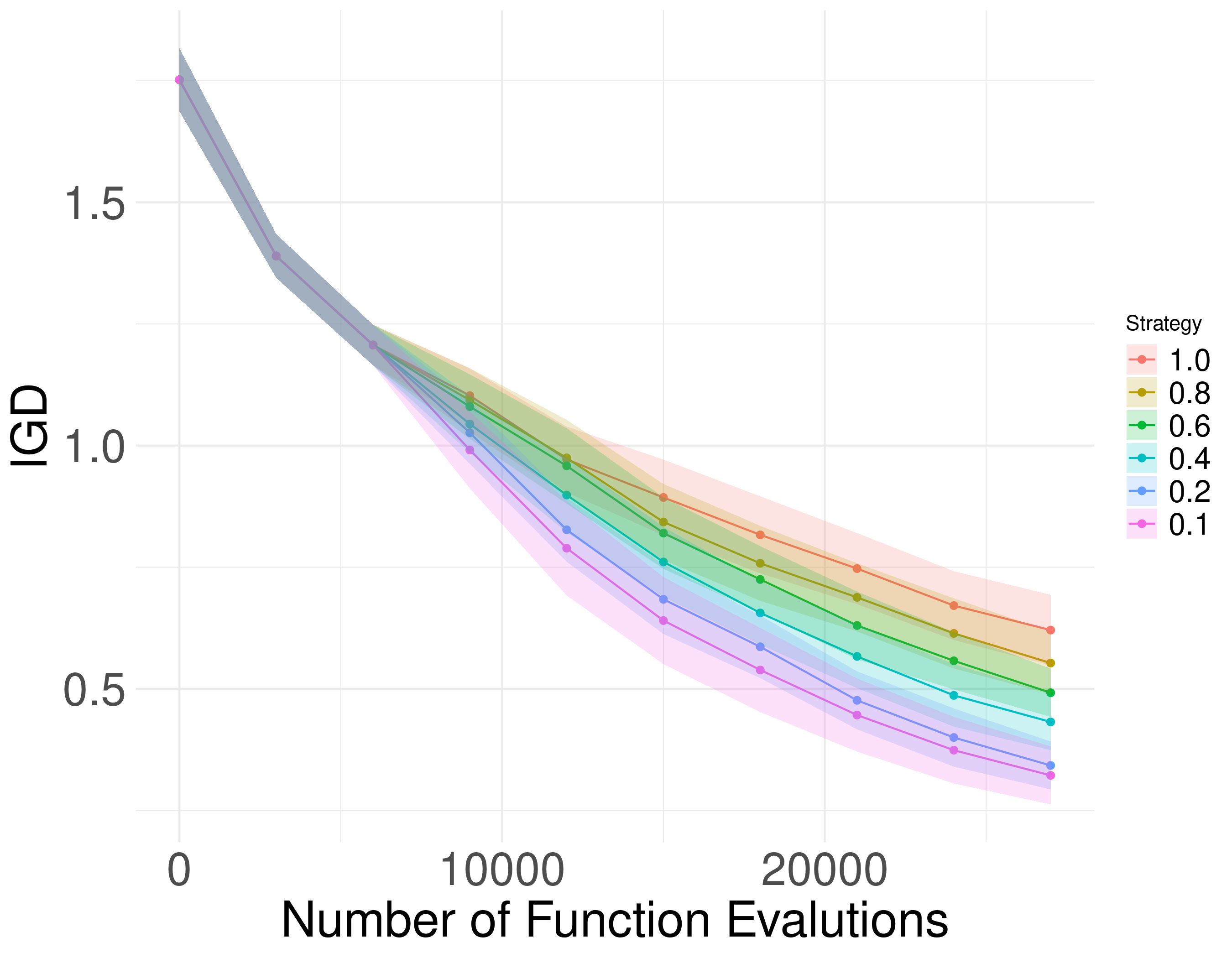}
        	\caption{UF6.}
	\end{subfigure}
	~~
    \begin{subfigure}[!t]{0.45\textwidth}
        	\includegraphics[width=1\textwidth]{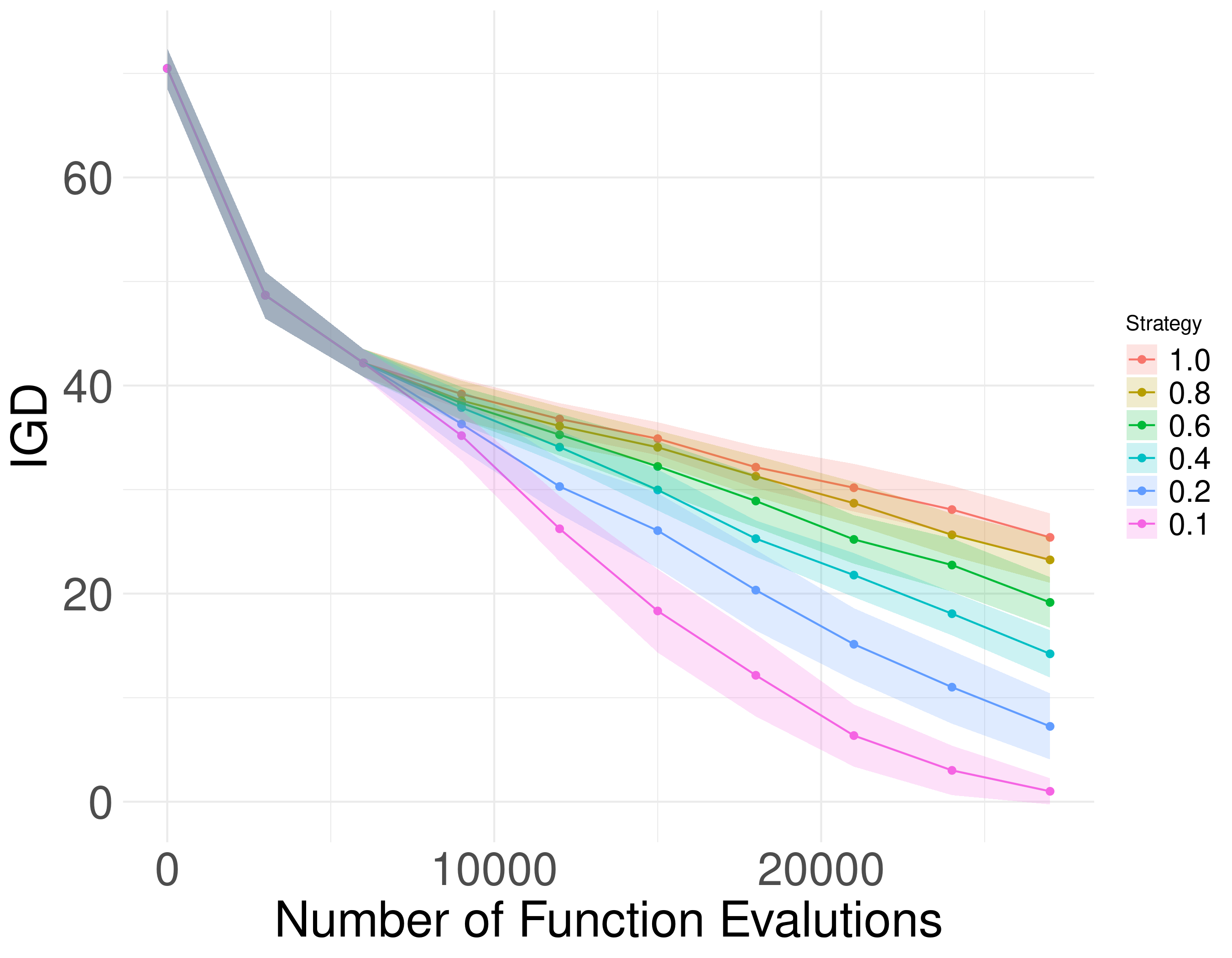}
        	\caption{DTLZ6.}
	\end{subfigure}
	\caption{Anytime IGD (lower is better) performance of MOEA/D-PS for different values of $ps$ on two functions.}
	\label{fig:anytime_igd}
\end{figure*}

Figures~\ref{fig:anytime_hv} illustrates the anytime performance of the MOEA/D with Partial Update Strategy in terms of hypervolume (higher is better) for two specific problems while Figure~\ref{fig:anytime_igd} illustrates the anytime performance of the MOEA/D with Partial Update Strategy in terms of IGD (lower is better). Please recall that all subproblems are selected regardless of $ps$ until iteration $\Delta t = 20$ (see Section~\ref{chap3:moead_ps_explanation} for details). Consistently with the end-of-run results, Figures~\ref{fig:anytime_hv} and~\ref{fig:anytime_igd} indicate that changing smaller percentages of the population at each iteration tends to result in better performance anytime during the search. While this is only illustrated here for two test problems, the same behavior is observed for almost all other problems. We consider these results as an indicative that smaller values of the $ps$ parameter result in faster and better convergence for the MOEA/D, at least for 2- and 3-objective problems with characteristics similar to the test ones employed in these experiments.

%% file: files/parameter_table.tex
\begin{table}[htb]
\centering
	\normalsize
	\caption{Experimental parameter settings.}
	\label{chap4:parameter_table}
    \begin{tabular}{|l|l|}
        \hline
        \rowcolor[gray]{.85}MOEA/D-DE parameters             & Value            \\ \hline
        DE mutation param.                                  & $F = 0.25$       \\ \hline
        \multirow{2}{*}{Polynomial mutation params.}        & $\eta_m = 20$    \\
                                                            & $p_m = 0.01$     \\ \hline
        Restricted Update param.                            & $nr = 2$         \\ \hline
        Locality parameter                                  & $\delta_p = 0.9$ \\ \hline
        Neighborhood size                                   & $T = 20$         \\ \hline
        \multirow{2}{*}{SLD decomposition param.}           & $h = 249$ (2 obj) \\
                                                            & $h = 25$ (3 obj) \\ \hline
        \multirow{2}{*}{Population size}                    & $N = 350$ (2 obj) \\
                                                            & $N = 351$ (3 obj) \\ \hline
        \multicolumn{2}{c}{}\\
        \hline
        \rowcolor[gray]{.85}Resource Allocation Parameters  & Value            \\ \hline
        Generations before RA starts                        & $\Delta t = 20$  \\ \hline

        \multicolumn{2}{c}{}\\
        \hline
        \rowcolor[gray]{.85}Experiment Parameters           & Value            \\ \hline
        Repeated runs                                       & 21               \\ \hline
        Computational budget                                & 30000 evals.            \\ \hline
\end{tabular}
\end{table}

%% file: images/box_UF6.tex
\begin{tikzpicture}[x=1pt,y=1pt]
\definecolor{fillColor}{RGB}{255,255,255}
\path[use as bounding box,fill=fillColor,fill opacity=0.00] (0,0) rectangle (216.81,271.01);
\begin{scope}
\path[clip] ( 57.88,150.85) rectangle (211.31,265.51);
\definecolor{drawColor}{gray}{0.92}

\path[draw=drawColor,line width= 0.3pt,line join=round] ( 57.88,150.93) --
	(211.31,150.93);

\path[draw=drawColor,line width= 0.3pt,line join=round] ( 57.88,172.41) --
	(211.31,172.41);

\path[draw=drawColor,line width= 0.3pt,line join=round] ( 57.88,193.89) --
	(211.31,193.89);

\path[draw=drawColor,line width= 0.3pt,line join=round] ( 57.88,215.38) --
	(211.31,215.38);

\path[draw=drawColor,line width= 0.3pt,line join=round] ( 57.88,236.86) --
	(211.31,236.86);

\path[draw=drawColor,line width= 0.3pt,line join=round] ( 57.88,258.34) --
	(211.31,258.34);

\path[draw=drawColor,line width= 0.3pt,line join=round] ( 73.79,150.85) --
	( 73.79,265.51);

\path[draw=drawColor,line width= 0.3pt,line join=round] (109.56,150.85) --
	(109.56,265.51);

\path[draw=drawColor,line width= 0.3pt,line join=round] (145.32,150.85) --
	(145.32,265.51);

\path[draw=drawColor,line width= 0.3pt,line join=round] (181.09,150.85) --
	(181.09,265.51);

\path[draw=drawColor,line width= 0.6pt,line join=round] ( 57.88,161.67) --
	(211.31,161.67);

\path[draw=drawColor,line width= 0.6pt,line join=round] ( 57.88,183.15) --
	(211.31,183.15);

\path[draw=drawColor,line width= 0.6pt,line join=round] ( 57.88,204.64) --
	(211.31,204.64);

\path[draw=drawColor,line width= 0.6pt,line join=round] ( 57.88,226.12) --
	(211.31,226.12);

\path[draw=drawColor,line width= 0.6pt,line join=round] ( 57.88,247.60) --
	(211.31,247.60);

\path[draw=drawColor,line width= 0.6pt,line join=round] ( 91.67,150.85) --
	( 91.67,265.51);

\path[draw=drawColor,line width= 0.6pt,line join=round] (127.44,150.85) --
	(127.44,265.51);

\path[draw=drawColor,line width= 0.6pt,line join=round] (163.21,150.85) --
	(163.21,265.51);

\path[draw=drawColor,line width= 0.6pt,line join=round] (198.97,150.85) --
	(198.97,265.51);
\definecolor{drawColor}{RGB}{51,51,51}
\definecolor{fillColor}{RGB}{51,51,51}

\path[draw=drawColor,draw opacity=0.50,line width= 0.4pt,line join=round,line cap=round,fill=fillColor,fill opacity=0.50] ( 70.22,213.35) circle (  1.96);

\path[draw=drawColor,draw opacity=0.50,line width= 0.4pt,line join=round,line cap=round,fill=fillColor,fill opacity=0.50] ( 70.22,156.06) circle (  1.96);
\definecolor{drawColor}{gray}{0.20}

\path[draw=drawColor,line width= 0.6pt,line join=round] ( 70.22,182.78) -- ( 70.22,194.04);

\path[draw=drawColor,line width= 0.6pt,line join=round] ( 70.22,174.09) -- ( 70.22,169.53);
\definecolor{fillColor}{RGB}{248,118,109}

\path[draw=drawColor,line width= 0.6pt,line join=round,line cap=round,fill=fillColor,fill opacity=0.50] ( 64.85,182.78) --
	( 64.85,174.09) --
	( 75.58,174.09) --
	( 75.58,182.78) --
	( 64.85,182.78) --
	cycle;

\path[draw=drawColor,line width= 1.1pt,line join=round] ( 64.85,179.58) -- ( 75.58,179.58);

\path[draw=drawColor,line width= 0.6pt,line join=round] ( 84.52,191.22) -- ( 84.52,208.01);

\path[draw=drawColor,line width= 0.6pt,line join=round] ( 84.52,175.09) -- ( 84.52,169.22);
\definecolor{fillColor}{RGB}{183,159,0}

\path[draw=drawColor,line width= 0.6pt,line join=round,line cap=round,fill=fillColor,fill opacity=0.50] ( 79.16,191.22) --
	( 79.16,175.09) --
	( 89.89,175.09) --
	( 89.89,191.22) --
	( 79.16,191.22) --
	cycle;

\path[draw=drawColor,line width= 1.1pt,line join=round] ( 79.16,182.18) -- ( 89.89,182.18);

\path[draw=drawColor,line width= 0.6pt,line join=round] (113.13,208.02) -- (113.13,225.47);

\path[draw=drawColor,line width= 0.6pt,line join=round] (113.13,194.95) -- (113.13,178.18);
\definecolor{fillColor}{RGB}{0,186,56}

\path[draw=drawColor,line width= 0.6pt,line join=round,line cap=round,fill=fillColor,fill opacity=0.50] (107.77,208.02) --
	(107.77,194.95) --
	(118.50,194.95) --
	(118.50,208.02) --
	(107.77,208.02) --
	cycle;

\path[draw=drawColor,line width= 1.1pt,line join=round] (107.77,200.49) -- (118.50,200.49);
\definecolor{drawColor}{RGB}{51,51,51}
\definecolor{fillColor}{RGB}{51,51,51}

\path[draw=drawColor,draw opacity=0.50,line width= 0.4pt,line join=round,line cap=round,fill=fillColor,fill opacity=0.50] (141.75,188.07) circle (  1.96);
\definecolor{drawColor}{gray}{0.20}

\path[draw=drawColor,line width= 0.6pt,line join=round] (141.75,217.93) -- (141.75,226.19);

\path[draw=drawColor,line width= 0.6pt,line join=round] (141.75,208.43) -- (141.75,196.83);
\definecolor{fillColor}{RGB}{0,191,196}

\path[draw=drawColor,line width= 0.6pt,line join=round,line cap=round,fill=fillColor,fill opacity=0.50] (136.38,217.93) --
	(136.38,208.43) --
	(147.11,208.43) --
	(147.11,217.93) --
	(136.38,217.93) --
	cycle;

\path[draw=drawColor,line width= 1.1pt,line join=round] (136.38,212.26) -- (147.11,212.26);

\path[draw=drawColor,line width= 0.6pt,line join=round] (170.36,235.39) -- (170.36,240.41);

\path[draw=drawColor,line width= 0.6pt,line join=round] (170.36,212.09) -- (170.36,196.81);
\definecolor{fillColor}{RGB}{97,156,255}

\path[draw=drawColor,line width= 0.6pt,line join=round,line cap=round,fill=fillColor,fill opacity=0.50] (164.99,235.39) --
	(164.99,212.09) --
	(175.72,212.09) --
	(175.72,235.39) --
	(164.99,235.39) --
	cycle;

\path[draw=drawColor,line width= 1.1pt,line join=round] (164.99,226.69) -- (175.72,226.69);

\path[draw=drawColor,line width= 0.6pt,line join=round] (198.97,249.48) -- (198.97,260.30);

\path[draw=drawColor,line width= 0.6pt,line join=round] (198.97,228.79) -- (198.97,209.64);
\definecolor{fillColor}{RGB}{245,100,227}

\path[draw=drawColor,line width= 0.6pt,line join=round,line cap=round,fill=fillColor,fill opacity=0.50] (193.61,249.48) --
	(193.61,228.79) --
	(204.34,228.79) --
	(204.34,249.48) --
	(193.61,249.48) --
	cycle;

\path[draw=drawColor,line width= 1.1pt,line join=round] (193.61,243.14) -- (204.34,243.14);
\end{scope}
\begin{scope}
\path[clip] ( 57.88, 30.69) rectangle (211.31,145.35);
\definecolor{drawColor}{gray}{0.92}

\path[draw=drawColor,line width= 0.3pt,line join=round] ( 57.88, 50.03) --
	(211.31, 50.03);

\path[draw=drawColor,line width= 0.3pt,line join=round] ( 57.88, 73.25) --
	(211.31, 73.25);

\path[draw=drawColor,line width= 0.3pt,line join=round] ( 57.88, 96.48) --
	(211.31, 96.48);

\path[draw=drawColor,line width= 0.3pt,line join=round] ( 57.88,119.70) --
	(211.31,119.70);

\path[draw=drawColor,line width= 0.3pt,line join=round] ( 57.88,142.92) --
	(211.31,142.92);

\path[draw=drawColor,line width= 0.3pt,line join=round] ( 73.79, 30.69) --
	( 73.79,145.35);

\path[draw=drawColor,line width= 0.3pt,line join=round] (109.56, 30.69) --
	(109.56,145.35);

\path[draw=drawColor,line width= 0.3pt,line join=round] (145.32, 30.69) --
	(145.32,145.35);

\path[draw=drawColor,line width= 0.3pt,line join=round] (181.09, 30.69) --
	(181.09,145.35);

\path[draw=drawColor,line width= 0.6pt,line join=round] ( 57.88, 38.42) --
	(211.31, 38.42);

\path[draw=drawColor,line width= 0.6pt,line join=round] ( 57.88, 61.64) --
	(211.31, 61.64);

\path[draw=drawColor,line width= 0.6pt,line join=round] ( 57.88, 84.86) --
	(211.31, 84.86);

\path[draw=drawColor,line width= 0.6pt,line join=round] ( 57.88,108.09) --
	(211.31,108.09);

\path[draw=drawColor,line width= 0.6pt,line join=round] ( 57.88,131.31) --
	(211.31,131.31);

\path[draw=drawColor,line width= 0.6pt,line join=round] ( 91.67, 30.69) --
	( 91.67,145.35);

\path[draw=drawColor,line width= 0.6pt,line join=round] (127.44, 30.69) --
	(127.44,145.35);

\path[draw=drawColor,line width= 0.6pt,line join=round] (163.21, 30.69) --
	(163.21,145.35);

\path[draw=drawColor,line width= 0.6pt,line join=round] (198.97, 30.69) --
	(198.97,145.35);
\definecolor{drawColor}{RGB}{51,51,51}
\definecolor{fillColor}{RGB}{51,51,51}

\path[draw=drawColor,draw opacity=0.50,line width= 0.4pt,line join=round,line cap=round,fill=fillColor,fill opacity=0.50] ( 70.22, 78.43) circle (  1.96);

\path[draw=drawColor,draw opacity=0.50,line width= 0.4pt,line join=round,line cap=round,fill=fillColor,fill opacity=0.50] ( 70.22,140.14) circle (  1.96);
\definecolor{drawColor}{gray}{0.20}

\path[draw=drawColor,line width= 0.6pt,line join=round] ( 70.22,121.12) -- ( 70.22,135.63);

\path[draw=drawColor,line width= 0.6pt,line join=round] ( 70.22,109.19) -- ( 70.22, 92.05);
\definecolor{fillColor}{RGB}{248,118,109}

\path[draw=drawColor,line width= 0.6pt,line join=round,line cap=round,fill=fillColor,fill opacity=0.50] ( 64.85,121.12) --
	( 64.85,109.19) --
	( 75.58,109.19) --
	( 75.58,121.12) --
	( 64.85,121.12) --
	cycle;

\path[draw=drawColor,line width= 1.1pt,line join=round] ( 64.85,115.41) -- ( 75.58,115.41);

\path[draw=drawColor,line width= 0.6pt,line join=round] ( 84.52,122.16) -- ( 84.52,129.78);

\path[draw=drawColor,line width= 0.6pt,line join=round] ( 84.52,102.07) -- ( 84.52, 77.48);
\definecolor{fillColor}{RGB}{183,159,0}

\path[draw=drawColor,line width= 0.6pt,line join=round,line cap=round,fill=fillColor,fill opacity=0.50] ( 79.16,122.16) --
	( 79.16,102.07) --
	( 89.89,102.07) --
	( 89.89,122.16) --
	( 79.16,122.16) --
	cycle;

\path[draw=drawColor,line width= 1.1pt,line join=round] ( 79.16,114.13) -- ( 89.89,114.13);

\path[draw=drawColor,line width= 0.6pt,line join=round] (113.13,106.50) -- (113.13,124.68);

\path[draw=drawColor,line width= 0.6pt,line join=round] (113.13, 85.10) -- (113.13, 75.08);
\definecolor{fillColor}{RGB}{0,186,56}

\path[draw=drawColor,line width= 0.6pt,line join=round,line cap=round,fill=fillColor,fill opacity=0.50] (107.77,106.50) --
	(107.77, 85.10) --
	(118.50, 85.10) --
	(118.50,106.50) --
	(107.77,106.50) --
	cycle;

\path[draw=drawColor,line width= 1.1pt,line join=round] (107.77, 95.23) -- (118.50, 95.23);

\path[draw=drawColor,line width= 0.6pt,line join=round] (141.75, 96.33) -- (141.75,112.03);

\path[draw=drawColor,line width= 0.6pt,line join=round] (141.75, 74.32) -- (141.75, 64.81);
\definecolor{fillColor}{RGB}{0,191,196}

\path[draw=drawColor,line width= 0.6pt,line join=round,line cap=round,fill=fillColor,fill opacity=0.50] (136.38, 96.33) --
	(136.38, 74.32) --
	(147.11, 74.32) --
	(147.11, 96.33) --
	(136.38, 96.33) --
	cycle;

\path[draw=drawColor,line width= 1.1pt,line join=round] (136.38, 84.49) -- (147.11, 84.49);

\path[draw=drawColor,line width= 0.6pt,line join=round] (170.36, 91.66) -- (170.36,109.52);

\path[draw=drawColor,line width= 0.6pt,line join=round] (170.36, 66.69) -- (170.36, 44.95);
\definecolor{fillColor}{RGB}{97,156,255}

\path[draw=drawColor,line width= 0.6pt,line join=round,line cap=round,fill=fillColor,fill opacity=0.50] (164.99, 91.66) --
	(164.99, 66.69) --
	(175.72, 66.69) --
	(175.72, 91.66) --
	(164.99, 91.66) --
	cycle;

\path[draw=drawColor,line width= 1.1pt,line join=round] (164.99, 72.79) -- (175.72, 72.79);

\path[draw=drawColor,line width= 0.6pt,line join=round] (198.97, 73.96) -- (198.97, 95.68);

\path[draw=drawColor,line width= 0.6pt,line join=round] (198.97, 53.14) -- (198.97, 35.90);
\definecolor{fillColor}{RGB}{245,100,227}

\path[draw=drawColor,line width= 0.6pt,line join=round,line cap=round,fill=fillColor,fill opacity=0.50] (193.61, 73.96) --
	(193.61, 53.14) --
	(204.34, 53.14) --
	(204.34, 73.96) --
	(193.61, 73.96) --
	cycle;

\path[draw=drawColor,line width= 1.1pt,line join=round] (193.61, 55.76) -- (204.34, 55.76);
\end{scope}
\begin{scope}
\path[clip] ( 17.96,150.85) rectangle ( 34.54,265.51);
\definecolor{drawColor}{gray}{0.10}

\node[text=drawColor,rotate= 90.00,anchor=base,inner sep=0pt, outer sep=0pt, scale=  0.88] at ( 29.28,208.18) {IGD};
\end{scope}
\begin{scope}
\path[clip] ( 17.96, 30.69) rectangle ( 34.54,145.35);
\definecolor{drawColor}{gray}{0.10}

\node[text=drawColor,rotate= 90.00,anchor=base,inner sep=0pt, outer sep=0pt, scale=  0.88] at ( 29.28, 88.02) {HV};
\end{scope}
\begin{scope}
\path[clip] (  0.00,  0.00) rectangle (216.81,271.01);
\definecolor{drawColor}{gray}{0.30}

\node[text=drawColor,anchor=base,inner sep=0pt, outer sep=0pt, scale=  0.88] at ( 91.67, 19.68) {0.25};

\node[text=drawColor,anchor=base,inner sep=0pt, outer sep=0pt, scale=  0.88] at (127.44, 19.68) {0.50};

\node[text=drawColor,anchor=base,inner sep=0pt, outer sep=0pt, scale=  0.88] at (163.21, 19.68) {0.75};

\node[text=drawColor,anchor=base,inner sep=0pt, outer sep=0pt, scale=  0.88] at (198.97, 19.68) {1.00};
\end{scope}
\begin{scope}
\path[clip] (  0.00,  0.00) rectangle (216.81,271.01);
\definecolor{drawColor}{gray}{0.30}

\node[text=drawColor,anchor=base east,inner sep=0pt, outer sep=0pt, scale=  0.88] at ( 52.93,158.64) {0.2};

\node[text=drawColor,anchor=base east,inner sep=0pt, outer sep=0pt, scale=  0.88] at ( 52.93,180.12) {0.3};

\node[text=drawColor,anchor=base east,inner sep=0pt, outer sep=0pt, scale=  0.88] at ( 52.93,201.61) {0.4};

\node[text=drawColor,anchor=base east,inner sep=0pt, outer sep=0pt, scale=  0.88] at ( 52.93,223.09) {0.5};

\node[text=drawColor,anchor=base east,inner sep=0pt, outer sep=0pt, scale=  0.88] at ( 52.93,244.57) {0.6};
\end{scope}
\begin{scope}
\path[clip] (  0.00,  0.00) rectangle (216.81,271.01);
\definecolor{drawColor}{gray}{0.30}

\node[text=drawColor,anchor=base east,inner sep=0pt, outer sep=0pt, scale=  0.88] at ( 52.93, 35.39) {0.65};

\node[text=drawColor,anchor=base east,inner sep=0pt, outer sep=0pt, scale=  0.88] at ( 52.93, 58.61) {0.70};

\node[text=drawColor,anchor=base east,inner sep=0pt, outer sep=0pt, scale=  0.88] at ( 52.93, 81.83) {0.75};

\node[text=drawColor,anchor=base east,inner sep=0pt, outer sep=0pt, scale=  0.88] at ( 52.93,105.06) {0.80};

\node[text=drawColor,anchor=base east,inner sep=0pt, outer sep=0pt, scale=  0.88] at ( 52.93,128.28) {0.85};
\end{scope}
\begin{scope}
\path[clip] (  0.00,  0.00) rectangle (216.81,271.01);
\definecolor{drawColor}{RGB}{0,0,0}

\node[text=drawColor,anchor=base,inner sep=0pt, outer sep=0pt, scale=  1.10] at (134.59,  7.64) {$ps$};
\end{scope}
\end{tikzpicture}

%% file: images/box_DTLZ6.tex
\begin{tikzpicture}[x=1pt,y=1pt]
\definecolor{fillColor}{RGB}{255,255,255}
\path[use as bounding box,fill=fillColor,fill opacity=0.00] (0,0) rectangle (216.81,271.01);
\begin{scope}
\path[clip] ( 53.48,150.85) rectangle (211.31,265.51);
\definecolor{drawColor}{gray}{0.92}

\path[draw=drawColor,line width= 0.3pt,line join=round] ( 53.48,173.74) --
	(211.31,173.74);

\path[draw=drawColor,line width= 0.3pt,line join=round] ( 53.48,209.13) --
	(211.31,209.13);

\path[draw=drawColor,line width= 0.3pt,line join=round] ( 53.48,244.52) --
	(211.31,244.52);

\path[draw=drawColor,line width= 0.3pt,line join=round] ( 69.85,150.85) --
	( 69.85,265.51);

\path[draw=drawColor,line width= 0.3pt,line join=round] (106.64,150.85) --
	(106.64,265.51);

\path[draw=drawColor,line width= 0.3pt,line join=round] (143.43,150.85) --
	(143.43,265.51);

\path[draw=drawColor,line width= 0.3pt,line join=round] (180.22,150.85) --
	(180.22,265.51);

\path[draw=drawColor,line width= 0.6pt,line join=round] ( 53.48,156.05) --
	(211.31,156.05);

\path[draw=drawColor,line width= 0.6pt,line join=round] ( 53.48,191.43) --
	(211.31,191.43);

\path[draw=drawColor,line width= 0.6pt,line join=round] ( 53.48,226.82) --
	(211.31,226.82);

\path[draw=drawColor,line width= 0.6pt,line join=round] ( 53.48,262.21) --
	(211.31,262.21);

\path[draw=drawColor,line width= 0.6pt,line join=round] ( 88.24,150.85) --
	( 88.24,265.51);

\path[draw=drawColor,line width= 0.6pt,line join=round] (125.04,150.85) --
	(125.04,265.51);

\path[draw=drawColor,line width= 0.6pt,line join=round] (161.83,150.85) --
	(161.83,265.51);

\path[draw=drawColor,line width= 0.6pt,line join=round] (198.62,150.85) --
	(198.62,265.51);
\definecolor{drawColor}{RGB}{51,51,51}
\definecolor{fillColor}{RGB}{51,51,51}

\path[draw=drawColor,draw opacity=0.50,line width= 0.4pt,line join=round,line cap=round,fill=fillColor,fill opacity=0.50] ( 66.17,164.66) circle (  1.96);
\definecolor{drawColor}{gray}{0.20}

\path[draw=drawColor,line width= 0.6pt,line join=round] ( 66.17,158.29) -- ( 66.17,159.72);

\path[draw=drawColor,line width= 0.6pt,line join=round] ( 66.17,156.06) -- ( 66.17,156.06);
\definecolor{fillColor}{RGB}{248,118,109}

\path[draw=drawColor,line width= 0.6pt,line join=round,line cap=round,fill=fillColor,fill opacity=0.50] ( 60.65,158.29) --
	( 60.65,156.06) --
	( 71.69,156.06) --
	( 71.69,158.29) --
	( 60.65,158.29) --
	cycle;

\path[draw=drawColor,line width= 1.1pt,line join=round] ( 60.65,156.08) -- ( 71.69,156.08);

\path[draw=drawColor,line width= 0.6pt,line join=round] ( 80.89,181.09) -- ( 80.89,196.07);

\path[draw=drawColor,line width= 0.6pt,line join=round] ( 80.89,167.04) -- ( 80.89,159.47);
\definecolor{fillColor}{RGB}{183,159,0}

\path[draw=drawColor,line width= 0.6pt,line join=round,line cap=round,fill=fillColor,fill opacity=0.50] ( 75.37,181.09) --
	( 75.37,167.04) --
	( 86.41,167.04) --
	( 86.41,181.09) --
	( 75.37,181.09) --
	cycle;

\path[draw=drawColor,line width= 1.1pt,line join=round] ( 75.37,175.04) -- ( 86.41,175.04);
\definecolor{drawColor}{RGB}{51,51,51}
\definecolor{fillColor}{RGB}{51,51,51}

\path[draw=drawColor,draw opacity=0.50,line width= 0.4pt,line join=round,line cap=round,fill=fillColor,fill opacity=0.50] (110.32,182.12) circle (  1.96);
\definecolor{drawColor}{gray}{0.20}

\path[draw=drawColor,line width= 0.6pt,line join=round] (110.32,203.41) -- (110.32,210.04);

\path[draw=drawColor,line width= 0.6pt,line join=round] (110.32,195.08) -- (110.32,185.64);
\definecolor{fillColor}{RGB}{0,186,56}

\path[draw=drawColor,line width= 0.6pt,line join=round,line cap=round,fill=fillColor,fill opacity=0.50] (104.80,203.41) --
	(104.80,195.08) --
	(115.84,195.08) --
	(115.84,203.41) --
	(104.80,203.41) --
	cycle;

\path[draw=drawColor,line width= 1.1pt,line join=round] (104.80,199.38) -- (115.84,199.38);

\path[draw=drawColor,line width= 0.6pt,line join=round] (139.75,221.07) -- (139.75,230.66);

\path[draw=drawColor,line width= 0.6pt,line join=round] (139.75,213.63) -- (139.75,202.97);
\definecolor{fillColor}{RGB}{0,191,196}

\path[draw=drawColor,line width= 0.6pt,line join=round,line cap=round,fill=fillColor,fill opacity=0.50] (134.23,221.07) --
	(134.23,213.63) --
	(145.27,213.63) --
	(145.27,221.07) --
	(134.23,221.07) --
	cycle;

\path[draw=drawColor,line width= 1.1pt,line join=round] (134.23,217.13) -- (145.27,217.13);

\path[draw=drawColor,line width= 0.6pt,line join=round] (169.18,234.21) -- (169.18,245.12);

\path[draw=drawColor,line width= 0.6pt,line join=round] (169.18,224.29) -- (169.18,215.38);
\definecolor{fillColor}{RGB}{97,156,255}

\path[draw=drawColor,line width= 0.6pt,line join=round,line cap=round,fill=fillColor,fill opacity=0.50] (163.67,234.21) --
	(163.67,224.29) --
	(174.70,224.29) --
	(174.70,234.21) --
	(163.67,234.21) --
	cycle;

\path[draw=drawColor,line width= 1.1pt,line join=round] (163.67,229.69) -- (174.70,229.69);
\definecolor{drawColor}{RGB}{51,51,51}
\definecolor{fillColor}{RGB}{51,51,51}

\path[draw=drawColor,draw opacity=0.50,line width= 0.4pt,line join=round,line cap=round,fill=fillColor,fill opacity=0.50] (198.62,260.30) circle (  1.96);

\path[draw=drawColor,draw opacity=0.50,line width= 0.4pt,line join=round,line cap=round,fill=fillColor,fill opacity=0.50] (198.62,257.14) circle (  1.96);
\definecolor{drawColor}{gray}{0.20}

\path[draw=drawColor,line width= 0.6pt,line join=round] (198.62,243.60) -- (198.62,247.95);

\path[draw=drawColor,line width= 0.6pt,line join=round] (198.62,235.32) -- (198.62,224.16);
\definecolor{fillColor}{RGB}{245,100,227}

\path[draw=drawColor,line width= 0.6pt,line join=round,line cap=round,fill=fillColor,fill opacity=0.50] (193.10,243.60) --
	(193.10,235.32) --
	(204.14,235.32) --
	(204.14,243.60) --
	(193.10,243.60) --
	cycle;

\path[draw=drawColor,line width= 1.1pt,line join=round] (193.10,239.81) -- (204.14,239.81);
\end{scope}
\begin{scope}
\path[clip] ( 53.48, 30.69) rectangle (211.31,145.35);
\definecolor{drawColor}{gray}{0.92}

\path[draw=drawColor,line width= 0.3pt,line join=round] ( 53.48, 31.03) --
	(211.31, 31.03);

\path[draw=drawColor,line width= 0.3pt,line join=round] ( 53.48, 55.29) --
	(211.31, 55.29);

\path[draw=drawColor,line width= 0.3pt,line join=round] ( 53.48, 79.56) --
	(211.31, 79.56);

\path[draw=drawColor,line width= 0.3pt,line join=round] ( 53.48,103.83) --
	(211.31,103.83);

\path[draw=drawColor,line width= 0.3pt,line join=round] ( 53.48,128.10) --
	(211.31,128.10);

\path[draw=drawColor,line width= 0.3pt,line join=round] ( 69.85, 30.69) --
	( 69.85,145.35);

\path[draw=drawColor,line width= 0.3pt,line join=round] (106.64, 30.69) --
	(106.64,145.35);

\path[draw=drawColor,line width= 0.3pt,line join=round] (143.43, 30.69) --
	(143.43,145.35);

\path[draw=drawColor,line width= 0.3pt,line join=round] (180.22, 30.69) --
	(180.22,145.35);

\path[draw=drawColor,line width= 0.6pt,line join=round] ( 53.48, 43.16) --
	(211.31, 43.16);

\path[draw=drawColor,line width= 0.6pt,line join=round] ( 53.48, 67.43) --
	(211.31, 67.43);

\path[draw=drawColor,line width= 0.6pt,line join=round] ( 53.48, 91.70) --
	(211.31, 91.70);

\path[draw=drawColor,line width= 0.6pt,line join=round] ( 53.48,115.97) --
	(211.31,115.97);

\path[draw=drawColor,line width= 0.6pt,line join=round] ( 53.48,140.23) --
	(211.31,140.23);

\path[draw=drawColor,line width= 0.6pt,line join=round] ( 88.24, 30.69) --
	( 88.24,145.35);

\path[draw=drawColor,line width= 0.6pt,line join=round] (125.04, 30.69) --
	(125.04,145.35);

\path[draw=drawColor,line width= 0.6pt,line join=round] (161.83, 30.69) --
	(161.83,145.35);

\path[draw=drawColor,line width= 0.6pt,line join=round] (198.62, 30.69) --
	(198.62,145.35);
\definecolor{drawColor}{RGB}{51,51,51}
\definecolor{fillColor}{RGB}{51,51,51}

\path[draw=drawColor,draw opacity=0.50,line width= 0.4pt,line join=round,line cap=round,fill=fillColor,fill opacity=0.50] ( 66.17,139.78) circle (  1.96);

\path[draw=drawColor,draw opacity=0.50,line width= 0.4pt,line join=round,line cap=round,fill=fillColor,fill opacity=0.50] ( 66.17,139.73) circle (  1.96);

\path[draw=drawColor,draw opacity=0.50,line width= 0.4pt,line join=round,line cap=round,fill=fillColor,fill opacity=0.50] ( 66.17,138.84) circle (  1.96);
\definecolor{drawColor}{gray}{0.20}

\path[draw=drawColor,line width= 0.6pt,line join=round] ( 66.17,140.14) -- ( 66.17,140.14);

\path[draw=drawColor,line width= 0.6pt,line join=round] ( 66.17,140.00) -- ( 66.17,139.84);
\definecolor{fillColor}{RGB}{248,118,109}

\path[draw=drawColor,line width= 0.6pt,line join=round,line cap=round,fill=fillColor,fill opacity=0.50] ( 60.65,140.14) --
	( 60.65,140.00) --
	( 71.69,140.00) --
	( 71.69,140.14) --
	( 60.65,140.14) --
	cycle;

\path[draw=drawColor,line width= 1.1pt,line join=round] ( 60.65,140.14) -- ( 71.69,140.14);

\path[draw=drawColor,line width= 0.6pt,line join=round] ( 80.89,137.69) -- ( 80.89,139.86);

\path[draw=drawColor,line width= 0.6pt,line join=round] ( 80.89,130.18) -- ( 80.89,123.43);
\definecolor{fillColor}{RGB}{183,159,0}

\path[draw=drawColor,line width= 0.6pt,line join=round,line cap=round,fill=fillColor,fill opacity=0.50] ( 75.37,137.69) --
	( 75.37,130.18) --
	( 86.41,130.18) --
	( 86.41,137.69) --
	( 75.37,137.69) --
	cycle;

\path[draw=drawColor,line width= 1.1pt,line join=round] ( 75.37,135.50) -- ( 86.41,135.50);

\path[draw=drawColor,line width= 0.6pt,line join=round] (110.32,120.53) -- (110.32,132.02);

\path[draw=drawColor,line width= 0.6pt,line join=round] (110.32,112.47) -- (110.32,108.12);
\definecolor{fillColor}{RGB}{0,186,56}

\path[draw=drawColor,line width= 0.6pt,line join=round,line cap=round,fill=fillColor,fill opacity=0.50] (104.80,120.53) --
	(104.80,112.47) --
	(115.84,112.47) --
	(115.84,120.53) --
	(104.80,120.53) --
	cycle;

\path[draw=drawColor,line width= 1.1pt,line join=round] (104.80,117.74) -- (115.84,117.74);

\path[draw=drawColor,line width= 0.6pt,line join=round] (139.75,104.27) -- (139.75,115.27);

\path[draw=drawColor,line width= 0.6pt,line join=round] (139.75, 92.32) -- (139.75, 80.54);
\definecolor{fillColor}{RGB}{0,191,196}

\path[draw=drawColor,line width= 0.6pt,line join=round,line cap=round,fill=fillColor,fill opacity=0.50] (134.23,104.27) --
	(134.23, 92.32) --
	(145.27, 92.32) --
	(145.27,104.27) --
	(134.23,104.27) --
	cycle;

\path[draw=drawColor,line width= 1.1pt,line join=round] (134.23, 97.29) -- (145.27, 97.29);
\definecolor{drawColor}{RGB}{51,51,51}
\definecolor{fillColor}{RGB}{51,51,51}

\path[draw=drawColor,draw opacity=0.50,line width= 0.4pt,line join=round,line cap=round,fill=fillColor,fill opacity=0.50] (169.18, 60.47) circle (  1.96);
\definecolor{drawColor}{gray}{0.20}

\path[draw=drawColor,line width= 0.6pt,line join=round] (169.18, 84.52) -- (169.18, 93.57);

\path[draw=drawColor,line width= 0.6pt,line join=round] (169.18, 76.93) -- (169.18, 69.02);
\definecolor{fillColor}{RGB}{97,156,255}

\path[draw=drawColor,line width= 0.6pt,line join=round,line cap=round,fill=fillColor,fill opacity=0.50] (163.67, 84.52) --
	(163.67, 76.93) --
	(174.70, 76.93) --
	(174.70, 84.52) --
	(163.67, 84.52) --
	cycle;

\path[draw=drawColor,line width= 1.1pt,line join=round] (163.67, 83.17) -- (174.70, 83.17);

\path[draw=drawColor,line width= 0.6pt,line join=round] (198.62, 70.52) -- (198.62, 81.36);

\path[draw=drawColor,line width= 0.6pt,line join=round] (198.62, 55.14) -- (198.62, 35.90);
\definecolor{fillColor}{RGB}{245,100,227}

\path[draw=drawColor,line width= 0.6pt,line join=round,line cap=round,fill=fillColor,fill opacity=0.50] (193.10, 70.52) --
	(193.10, 55.14) --
	(204.14, 55.14) --
	(204.14, 70.52) --
	(193.10, 70.52) --
	cycle;

\path[draw=drawColor,line width= 1.1pt,line join=round] (193.10, 64.64) -- (204.14, 64.64);
\end{scope}
\begin{scope}
\path[clip] ( 17.96,150.85) rectangle ( 34.54,265.51);
\definecolor{drawColor}{gray}{0.10}

\node[text=drawColor,rotate= 90.00,anchor=base,inner sep=0pt, outer sep=0pt, scale=  0.88] at ( 29.28,208.18) {IGD};
\end{scope}
\begin{scope}
\path[clip] ( 17.96, 30.69) rectangle ( 34.54,145.35);
\definecolor{drawColor}{gray}{0.10}

\node[text=drawColor,rotate= 90.00,anchor=base,inner sep=0pt, outer sep=0pt, scale=  0.88] at ( 29.28, 88.02) {HV};
\end{scope}
\begin{scope}
\path[clip] (  0.00,  0.00) rectangle (216.81,271.01);
\definecolor{drawColor}{gray}{0.30}

\node[text=drawColor,anchor=base,inner sep=0pt, outer sep=0pt, scale=  0.88] at ( 88.24, 19.68) {0.25};

\node[text=drawColor,anchor=base,inner sep=0pt, outer sep=0pt, scale=  0.88] at (125.04, 19.68) {0.50};

\node[text=drawColor,anchor=base,inner sep=0pt, outer sep=0pt, scale=  0.88] at (161.83, 19.68) {0.75};

\node[text=drawColor,anchor=base,inner sep=0pt, outer sep=0pt, scale=  0.88] at (198.62, 19.68) {1.00};
\end{scope}
\begin{scope}
\path[clip] (  0.00,  0.00) rectangle (216.81,271.01);
\definecolor{drawColor}{gray}{0.30}

\node[text=drawColor,anchor=base east,inner sep=0pt, outer sep=0pt, scale=  0.88] at ( 48.53,153.02) {0};

\node[text=drawColor,anchor=base east,inner sep=0pt, outer sep=0pt, scale=  0.88] at ( 48.53,188.40) {10};

\node[text=drawColor,anchor=base east,inner sep=0pt, outer sep=0pt, scale=  0.88] at ( 48.53,223.79) {20};

\node[text=drawColor,anchor=base east,inner sep=0pt, outer sep=0pt, scale=  0.88] at ( 48.53,259.18) {30};
\end{scope}
\begin{scope}
\path[clip] (  0.00,  0.00) rectangle (216.81,271.01);
\definecolor{drawColor}{gray}{0.30}

\node[text=drawColor,anchor=base east,inner sep=0pt, outer sep=0pt, scale=  0.88] at ( 48.53, 40.13) {0.6};

\node[text=drawColor,anchor=base east,inner sep=0pt, outer sep=0pt, scale=  0.88] at ( 48.53, 64.40) {0.7};

\node[text=drawColor,anchor=base east,inner sep=0pt, outer sep=0pt, scale=  0.88] at ( 48.53, 88.67) {0.8};

\node[text=drawColor,anchor=base east,inner sep=0pt, outer sep=0pt, scale=  0.88] at ( 48.53,112.94) {0.9};

\node[text=drawColor,anchor=base east,inner sep=0pt, outer sep=0pt, scale=  0.88] at ( 48.53,137.20) {1.0};
\end{scope}
\begin{scope}
\path[clip] (  0.00,  0.00) rectangle (216.81,271.01);
\definecolor{drawColor}{RGB}{0,0,0}

\node[text=drawColor,anchor=base,inner sep=0pt, outer sep=0pt, scale=  1.10] at (132.39,  7.64) {$ps$};
\end{scope}
\end{tikzpicture}

%% file: files/stats.tex
\begin{table}[htbp]
	\centering
	\small
	\caption{Statistical significance of differences in median IGD and HV, associated with different $ps$ values. Values are Hommel-adjusted p-values of Wilcoxon Rank-sum tests. ``$\uparrow$" indicates superiority of the column method, and ``$\approx$" indicates differences not statistically significant ($95\%$ confidence level).}
	\label{chap4:pvals}
	\begin{tabular}{l|lllll}
		\hline
		\rowcolor[gray]{.8}\multicolumn{6}{c}{\textbf{IGD}}\\\hline
		\rowcolor[gray]{.95}\textbf{ps} & \textbf{10\%} & \textbf{20\%} & \textbf{40\%} & \textbf{60\%} & \textbf{80\%} \\ \hline
		\textbf{20\%}  & 0.098 $\approx$     &                    &                   &                   &        \\ \hline
		\textbf{40\%}  & 6.4e-4 $\uparrow$   & 0.006 $\uparrow$   &                   &                   &        \\ \hline
		\textbf{60\%}  & 1.8e-4 $\uparrow$   &  7.6e-5 $\uparrow$ & 7.6e-5 $\uparrow$ &                   &        \\ \hline
		\textbf{80\%}  & 7.6e-5 $\uparrow$   & 7.6e-5 $\uparrow$  & 7.6e-5 $\uparrow$ & 7.6e-5 $\uparrow$ &        \\ \hline
		\textbf{100\%} & 7.6e-5 $\uparrow$   & 7.6e-5 $\uparrow$  & 7.6e-5 $\uparrow$ & 7.6e-5 $\uparrow$ & 7.6e-5 $\uparrow$ \\
		\hline
		\multicolumn{6}{c}{}\\\hline
		\rowcolor[gray]{.8}\multicolumn{6}{c}{\textbf{HV}}\\\hline
		\rowcolor[gray]{.95}\textbf{ps} & \textbf{10\%} & \textbf{20\%} & \textbf{40\%} & \textbf{60\%} & \textbf{80\%} \\ \hline
		\textbf{20\%}  & 0.185 $\approx$     &                    &                   &                   &        \\ \hline
		\textbf{40\%}  & 0.002 $\uparrow$    & 9.6e-4 $\uparrow$  &                   &                   &        \\ \hline
		\textbf{60\%}  & 9.6e-4 $\uparrow$   & 9.6e-4 $\uparrow$  & 0.001 $\uparrow$  &                   &        \\ \hline
		\textbf{80\%}  & 9.6e-4 $\uparrow$   & 9.6e-4 $\uparrow$  & 9.6e-4 $\uparrow$ & 9.6e-4 $\uparrow$ &        \\ \hline
		\textbf{100\%} & 9.6e-4 $\uparrow$   & 9.6e-4 $\uparrow$  & 9.6e-4 $\uparrow$ & 9.6e-4 $\uparrow$ & 9.6e-4 $\uparrow$ \\
		\hline
	\end{tabular}
\end{table}

%% file: files/5_comparision_study.tex
\section{Comparison Study}
\label{section:comparison}

In the previous section we investigated the influence of different values of the control parameter $ps$ on the performance of the MOEA/D-PS. The results indicate that low $ps$ values are associated with (anytime) improvements in IGD and HV values. In this section we compare the MOEA/D-PS (using $ps = 0.1$) against the original MOEA/D-DE and a MOEA/D-DE with resource allocation based on RI\cite{zhang2009performance, zhou2016all}. The same test problems described in the previous section were used.

Table \ref{chap5:stats_all} tabulates the mean results obtained by the MOEA/D-PS with $\textit{ps} = 0.1$, the pure MOEA/D-DE and the MOEA/D-DE with RI-based resource allocation, for all test problems. It is clear that the MOEA/D-PS results are considerably better when compared to the other methods, not only in terms of IGD and HV, but also on the mean proportion of nondominated solutions (NDOM) that it returns in the final population. Table \ref{chap5:pvals} presents the results of statistical pairwise comparisons using the same methodology described in subsection~\ref{evaluation}, corroborating the results observed in Table \ref{chap5:stats_all}.

Looking at the proportion of non-dominated solutions (NDOM) in Table~\ref{chap5:stats_all}, we see that randomly updating a small fraction of the subproblems at each iteration resulted in the highest value on all functions, often with a substantial lead. 
In our view, a higher proportion of non-dominated solutions suggests a better, more diverse set of solutions in the objective space, which would indicate the use of the partial update strategy (under a low $ps$ value) as an interesting strategy for improving convergence (subsection~\ref{chap4:anytime}) and diversity in the MOEA/D.

\begin{table}[htbp]
\centering
\caption[Means and standard errors for IGD, HV and proportion of nondominated solutions.]{Means and standard errors for IGD, HV and proportion of nondominated solutions (NDOM), for each algorithm-problem pair. The best point estimate for each problem is highlighted.}
\label{chap5:stats_all}
\begin{tabular}{l|lll}
    \hline
    \rowcolor[gray]{.8}\multicolumn{4}{c}{\textbf{IGD}}\\\hline
    \rowcolor[gray]{.95} & MOEA/D-PS & MOEA/D-DE & RI \\ 
    \hline
    UF1 & $\mathbf{0.26\pm 0.002}$ & $0.55\pm 0.003$ & $0.37\pm 0.002$ \\ 
    UF2 & $0.1\pm 0.001$ & $0.12\pm 0.001$ & $\mathbf{0.096\pm 0.001}$ \\ 
    UF3 & $\mathbf{0.28\pm 0.001}$ & $0.31\pm 0.001$ & $0.29\pm 0.001$ \\ 
    UF4 & $\mathbf{0.11\pm 0.001}$ & $\mathbf{0.11\pm 0.001}$ & $\mathbf{0.11\pm 0.001}$ \\ 
    UF5 & $\mathbf{1.1\pm 0.005}$ & $1.7\pm 0.003$ & $1.3\pm 0.004$ \\ 
    UF6 & $\mathbf{0.29\pm 0.003}$ & $0.56\pm 0.003$ & $0.38\pm 0.003$ \\ 
    UF7 & $\mathbf{0.26\pm 0.002}$ & $0.53\pm 0.003$ & $0.36\pm 0.003$ \\ 
    UF8 & $\mathbf{0.27\pm 0.001}$ & $0.31\pm 0.001$ & $0.3\pm 0.001$ \\ 
    UF9 & $\mathbf{0.42\pm 0.001}$ & $0.47\pm 0.001$ & $0.46\pm 0.001$ \\ 
    UF10 & $\mathbf{2.1\pm 0.023}$ & $3.4\pm 0.008$ & $2.3\pm 0.012$ \\ 
    \hline
    DTLZ1 & $\mathbf{230\pm 5.5}$ & $440\pm 6.3$ & $260\pm 5.7$ \\ 
    DTLZ2 & $\mathbf{0.11\pm 0.001}$ & $0.18\pm 0.001$ & $0.13\pm 0.001$ \\ 
    DTLZ3 & $\mathbf{610\pm 16}$ & $970\pm 19$ & $660\pm 17$ \\ 
    DTLZ4 & $\mathbf{0.12\pm 0.002}$ & $0.23\pm 0.002$ & $0.18\pm 0.006$ \\ 
    DTLZ5 & $\mathbf{0.11\pm 0.001}$ & $0.19\pm 0.001$ & $0.13\pm 0.001$ \\ 
    DTLZ6 & $\mathbf{0.37\pm 0.03}$ & $24\pm 0.11$ & $13\pm 0.14$ \\ 
    DTLZ7 & $\mathbf{0.4\pm 0.013}$ & $3.9\pm 0.013$ & $1.9\pm 0.034$ \\ 
	\hline
	\multicolumn{4}{c}{}\\\hline
    \rowcolor[gray]{.8}\multicolumn{4}{c}{\textbf{HV}}\\\hline
    \rowcolor[gray]{.95} & MOEA/D-PS & MOEA/D-DE & RI \\ 
    \hline
    UF1 & $\mathbf{0.86\pm 0.001}$ & $0.74\pm 0.001$ & $0.82\pm 0.001$ \\ 
    UF2 & $0.79\pm 0.001$ & $0.76\pm 0.001$ & $\mathbf{0.81\pm 0.001}$ \\ 
    UF3 & $\mathbf{0.57\pm 0.003}$ & $0.52\pm 0.001$ & $0.55\pm 0.002$ \\ 
    UF4 & $\mathbf{0.37\pm 0.001}$ & $\mathbf{0.37\pm 0.001}$ & $\mathbf{0.37\pm 0.001}$ \\ 
    UF5 & $\mathbf{0.72\pm 0.002}$ & $0.57\pm 0.001$ & $0.69\pm 0.001$ \\ 
    UF6 & $\mathbf{0.81\pm 0.001}$ & $0.7\pm 0.002$ & $0.78\pm 0.001$ \\ 
    UF7 & $\mathbf{0.83\pm 0.001}$ & $0.71\pm 0.001$ & $0.79\pm 0.001$ \\ 
    UF8 & $\mathbf{0.85\pm 0.001}$ & $0.81\pm 0.001$ & $\mathbf{0.85\pm 0.001}$ \\ 
    UF9 & $\mathbf{0.78\pm 0.002}$ & $0.73\pm 0.001$ & $0.74\pm 0.001$ \\ 
    UF10 & $\mathbf{0.81\pm 0.003}$ & $0.67\pm 0.001$ & $0.8\pm 0.002$ \\ 
    \hline
    DTLZ1 & $\mathbf{1\pm 6.5e-5}$ & $0.99\pm 9.2e-5$ & $\mathbf{1\pm 6.9e-5}$ \\ 
    DTLZ2 & $\mathbf{0.92\pm 0.001}$ & $0.91\pm 0.001$ & $\mathbf{0.92\pm 0.001}$ \\ 
    DTLZ3 & $\mathbf{0.98\pm 0.001}$ & $0.96\pm 0.001$ & $\mathbf{0.98\pm 0.001}$ \\ 
    DTLZ4 & $\mathbf{0.98\pm 5.6e-5}$ & $0.97\pm 7.2e-5$ & $0.97\pm 0.001$ \\ 
    DTLZ5 & $\mathbf{0.92\pm 0.001}$ & $0.91\pm 0.001$ & $\mathbf{0.92\pm 0.001}$ \\ 
    DTLZ6 & $\mathbf{1\pm 5.8e-5}$ & $0.68\pm 0.002$ & $0.89\pm 0.001$ \\ 
    DTLZ7 & $\mathbf{0.88\pm 0.001}$ & $0.56\pm 0.001$ & $0.75\pm 0.002$ \\ \hline
    \multicolumn{4}{c}{}\\\hline
    \rowcolor[gray]{.8}\multicolumn{4}{c}{\textbf{NDOM}}\\\hline
    \rowcolor[gray]{.95} & MOEA/D-PS & MOEA/D-DE & RI \\ 
    \hline
    UF1 & $\mathbf{0.89\pm 0.003}$ & $0.27\pm 0.002$ & $0.45\pm 0.005$ \\ 
    UF2 & $\mathbf{0.96\pm 0.002}$ & $0.42\pm 0.003$ & $0.7\pm 0.008$ \\ 
    UF3 & $\mathbf{0.92\pm 0.002}$ & $0.23\pm 0.002$ & $0.43\pm 0.005$ \\ 
    UF4 & $\mathbf{0.9\pm 0.003}$ & $0.68\pm 0.003$ & $0.81\pm 0.004$ \\ 
    UF5 & $\mathbf{0.85\pm 0.005}$ & $0.19\pm 0.001$ & $0.43\pm 0.005$ \\ 
    UF6 & $\mathbf{0.86\pm 0.003}$ & $0.29\pm 0.002$ & $0.44\pm 0.004$ \\ 
    UF7 & $\mathbf{0.92\pm 0.002}$ & $0.31\pm 0.002$ & $0.5\pm 0.004$ \\ 
    UF8 & $\mathbf{0.99\pm 0.001}$ & $0.54\pm 0.004$ & $0.94\pm 0.002$ \\ 
    UF9 & $\mathbf{0.99\pm 0.001}$ & $0.55\pm 0.003$ & $0.88\pm 0.004$ \\ 
    UF10 & $\mathbf{0.94\pm 0.003}$ & $0.44\pm 0.004$ & $0.87\pm 0.004$ \\ 
    \hline
    DTLZ1 & $\mathbf{0.93\pm 0.004}$ & $0.1\pm 0.002$ & $0.51\pm 0.01$ \\ 
    DTLZ2 & $\mathbf{0.96\pm 0.002}$ & $0.3\pm 0.002$ & $0.69\pm 0.01$ \\ 
    DTLZ3 & $\mathbf{0.75\pm 0.01}$ & $0.046\pm 0.001$ & $0.19\pm 0.006$ \\ 
    DTLZ4 & $\mathbf{0.74\pm 0.006}$ & $0.18\pm 0.002$ & $0.51\pm 0.009$ \\ 
    DTLZ5 & $\mathbf{0.96\pm 0.002}$ & $0.3\pm 0.002$ & $0.69\pm 0.009$ \\ 
    DTLZ6 & $\mathbf{0.91\pm 0.009}$ & $0.063\pm 0.002$ & $0.15\pm 0.004$ \\ 
    DTLZ7 & $\mathbf{0.84\pm 0.005}$ & $0.22\pm 0.003$ & $0.45\pm 0.012$ \\ 
   \hline
\end{tabular}
\end{table}

\begin{table}[htbp]
	\centering
	\small
	\caption{Statistical significance of differences in median IGD, HV and NDOM, for the three algorithms tested in this section. Values are Hommel-adjusted p-values of Wilcoxon Rank-sum tests. ``$\uparrow$" indicates superiority of the column method, and ``$\approx$" indicates differences not statistically significant ($95\%$ confidence level).}
	\label{chap5:pvals}
	\begin{tabular}{l|ll}
		\hline
		\rowcolor[gray]{.8}\multicolumn{3}{c}{\textbf{IGD}}\\\hline
		\rowcolor[gray]{.95}\textbf{} & MOEA/D-PS & MOEA/D-RI \\ \hline
		MOEA/D-RI  & 2.1e-4 $\uparrow$   &                    \\ \hline
		MOEA/D-DE  & 3.1e-5 $\uparrow$   & 3.1e-5 $\uparrow$   \\ 
		\hline
		\multicolumn{3}{c}{}\\\hline
		\rowcolor[gray]{.8}\multicolumn{3}{c}{\textbf{HV}}\\\hline
		\rowcolor[gray]{.95}\textbf{} & MOEA/D-PS & MOEA/D-RI \\ \hline
		MOEA/D-RI  & 6.4e-4 $\uparrow$   &                    \\ \hline
		MOEA/D-DE  & 0.0025 $\uparrow$   & 6.4e-4 $\uparrow$   \\ 
		\hline
		\multicolumn{3}{c}{}\\\hline
		\rowcolor[gray]{.8}\multicolumn{3}{c}{\textbf{NDOM}}\\\hline
		\rowcolor[gray]{.95}\textbf{} & MOEA/D-PS & MOEA/D-RI \\ \hline
		MOEA/D-RI  & 3.2e-4 $\uparrow$   &                    \\ \hline
		MOEA/D-DE  & 3.2e-4 $\uparrow$   & 3.2e-4 $\uparrow$   \\ 
		\hline	
	\end{tabular}
\end{table}

%% file: files/6_conclusion.tex
\section{Conclusion}
\label{section:conclusion}

In this work we presented a random partial update strategy for the MOEA/D, which was incorporated into a simple algorithm (MOEA/D-PS). The partial update strategy adds one 
control parameter ($ps$), which regulates the proportion of the population that is 
selected for variation at any iteration. 

Six $ps$ values were investigated experimentally, revealing a strong association between more conservative updating of the MOEA/D population (i.e., lower $ps$ values) and improved performance. Based on these experiments we suggest using low $ps$ values, such as $ps = 0.1$, but more thorough sensitivity analyses should be conducted to refine our understanding of these effects.

In addition, we showed that the MOEA/D-PS with $ps=0.1$ values was able to outperform the pure MOEA/D-DE as well as a resource allocation MOEA/D based on the well-know RI priority function. This suggests that the MOEA/D benefits more from having slower population dynamics than from a specific prioritization of subproblems based on the relative improvement criteria. 

This study raises two issues that we consider important for further understanding the 
effect of Partial Update strategies. The first is whether MOEA/D-PS would benefit from 
adapting the $ps$ value throughout the search, either using a fixed schedule, or through 
online adaptation. The second are the interaction effects between the $ps$ value and 
other components of MOEA/D, such as decomposition strategy, neighborhood strategies, 
and other parameters of the algorithm. These would be interesting questions for further investigation.


